\documentclass{article}

\usepackage{arxiv}

\usepackage[utf8]{inputenc} % allow utf-8 input
\usepackage[T1]{fontenc}    % use 8-bit T1 fonts
\usepackage{hyperref}       % hyperlinks
\usepackage{url}            % simple URL typesetting
\usepackage{booktabs}       % professional-quality tables
\usepackage{amsfonts}       % blackboard math symbols
\usepackage{nicefrac}       % compact symbols for 1/2, etc.
\usepackage{microtype}      % microtypography
\usepackage{lipsum}
\usepackage{graphicx}
\usepackage{amssymb}
\usepackage{booktabs}
\usepackage[round, sort]{natbib}
\usepackage{subfig}
\graphicspath{ {./images/} }

\title{Long text outline generation: Chinese text outline based on unsupervised framework and large language model}

\author{
 Yan Yan  \\
  School of Management\\
  University of Melbourne\\
   \And
 Yuanchi Ma \\
  School of Coumputer\\
  Beijing Institute of Technology\\
}

\begin{document}
\maketitle
\begin{abstract}
Outline generation aims to reveal the internal structure of a document by identifying underlying chapter relationships and generating corresponding chapter summaries. Although existing deep learning methods and large models perform well on small- and medium-sized texts, they struggle to produce readable outlines for very long texts (such as fictional works), often failing to segment chapters coherently. In this paper, we propose a novel outline generation method for Chinese, combining an unsupervised framework with large models. Specifically, the method first generates chapter feature graph data based on entity and syntactic dependency relationships. Then, a representation module based on graph attention layers learns deep embeddings of the chapter graph data. Using these chapter embeddings, we design an operator based on Markov chain principles to segment plot boundaries. Finally, we employ a large model to generate summaries of each plot segment and produce the overall outline. We evaluate our model based on segmentation accuracy and outline readability, and our performance outperforms several deep learning models and large models in comparative evaluations.
\end{abstract}

\section{Introduction}

Well-written stories are often composed of numerous semantically coherent chapters, with each chapter or group of chapters centered around a specific theme. An outline can concisely capture the content structure of a document, providing clear guidance for navigation and significantly reducing the cognitive burden of understanding the entire text. Furthermore, it helps uncover the underlying thematic structure of the text. Outline generation captures various thematic elements of a text, including subtitles, plot points, and other key aspects of the narrative. Additionally, outlines facilitate a wide range of text analysis applications. They are not only beneficial for traditional downstream NLP tasks, such as document summarization~\cite{topic1} and discourse parsing~\cite{topic2}, but also play a crucial role in large language models (LLMs). For example, during retrieval-augmented generation (RAG) in large language models~\cite{LLM-RAG}, it is essential to extract the necessary information from long documents. The paragraph-level thematic structure of a document can aid in quickly locating the approximate position of the required content within a lengthy text, thereby reducing the search space. The relevant process concept is outlined as follows.

\begin{figure}[t]
	\centering
	\includegraphics[width=1\linewidth]{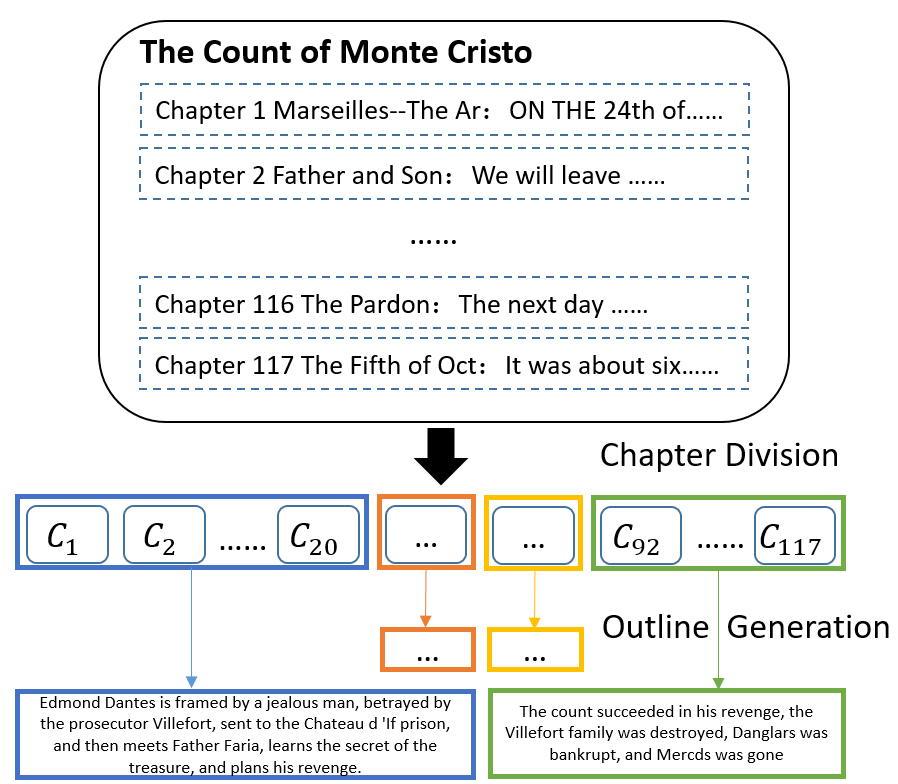}
	\caption{Multimodal
nonparametric clustering algorithm via Monte Carlo method and variational autoencoder diagram.}
	\label{flowchat}%文中引用该图片代号
\end{figure}

Previous work on outline generation~\cite{pr_outline1,pr_outline2} has primarily focused on short- and medium-length texts, such as news articles and announcements, helping readers quickly grasp the structure of the content. The vertical domains involved include sociology, psychology and economics~\cite{diff2}. These methods typically generate outlines based on natural paragraphs. However, for fictional works such as Game of Thrones (GoT), the Marvel Comics Universe (MCU), Greek mythology, or epic novels like Leo Tolstoy’s War and Peace or Don Winslow’s The Cartel, the task of outline generation is much more challenging due to their length and intricate semantic structures~\cite{diff1}. While novels are generally meant for entertainment, some of these works also reflect subcultural trends and capture the zeitgeist of particular eras. Analyzing their narrative structures and networks is of great interest to scholars in the humanities. For instance, War and Peace is set against the backdrop of Napoleon's wars in Russia, while The Cartel trilogy intertwines both fact and fiction concerning drug trafficking. Therefore, generating outlines for such long texts not only helps general readers quickly understand the relevant plots but also provides a valuable data foundation for historians, social scientists, media psychologists, and cultural studies scholars to conduct deeper analyses of these complex works~\cite{novel}.

Long fictional texts are composed of significant plot points, where the focus is less on basic relationships such as birthplaces or spouses, and more on specific events like alliances, betrayals, killings, or clan conflicts. These fictional works often have large fan communities, and search engines frequently receive queries like "When does Xiao Yan obtain the Bone Spirit Cold Flame?" (Battle Through the Heavens)~\cite{twitter}. For texts spanning millions of words, it is evidently challenging to locate a particular plot without an outline. Therefore, it is necessary to develop an outline generation method tailored to long fictional texts to address this problem. Upon further examination of the outline generation challenge, we observe that it actually involves two structured prediction tasks: (1) identifying chapter features and plot boundaries, and (2) generating chapter summary titles. These two tasks correspond to predicting the hierarchical relationship between chapters and summarizing individual chapters. While the second task can be well-handled by existing large language models (LLMs), particularly for short- and medium-length texts, LLMs often exhibit inaccuracies and increased hallucination issues when applied to longer texts. For instance, when tasked with summarizing a work of over a million words, LLMs may overlook key plot points, preventing readers from fully grasping the narrative [add experiment]. Therefore, we consider whether an ideal outline generation framework could extend the strengths of large models to ultra-long texts. The key challenge here lies in accurately identifying the sequence and features of chapters to obtain precise plot boundaries.

In our work, we propose a new end-to-end architecture to address this challenge. The key idea is to enhance large language model (LLM) outputs by guiding them with enriched information, specifically by determining plot boundaries through a neural network before using them to guide the LLM in generating a detailed outline. We posit that graph data can better represent relationships between entities within chapters, thus reflecting chapter characteristics more effectively. Therefore, our method first generates entity nodes through a chapter-level graph data generation module, followed by constructing the adjacency matrix between nodes based on syntactic dependency relationships. For node feature vectors, we not only select entity word vectors but also expand the feature set to include the tf-idf matrix of the entities, and we incorporate chapter numbers to represent contextual coherence. We then apply an improved graph neural network (GNN) based on graph attention layers (GAT) to learn from the chapter graph data. To this, we add a convolutional module for feature extraction and dimensionality reduction of deep chapter graph embeddings. For each chapter embedding, we perform chain-based prediction: specifically, we determine significant plot points and boundaries using Markov chains and path dependence based on their potential distances in feature space. Finally, LLMs are used to generate the themes and summaries for each plot segment, resulting in the final outline.

To facilitate our research, we constructed a new benchmark dataset in Chinese. As we observed, ultra-long texts in the domain of fictional literature often consist of millions of words. In this dataset, we not only provide the original literary works but also manually generated outlines to serve as a reference for evaluating experimental performance. For assessment, we compared several state-of-the-art methods to verify the effectiveness of our model, including rule-based models and large language models. The experimental results demonstrate that our proposed method significantly outperforms all baselines. We also conducted a detailed analysis of the proposed model, including an analysis of the readability of the generated outlines, to better understand the learned content structure.

The contributions of our work are summarized as follows:
\begin{itemize}
\item We developed a model for the task of outline generation for ultra-long documents, presenting a novel solution that combines unsupervised learning frameworks with large models.
\item  We established a public dataset for the outline generation (OG) task, which includes multiple ultra-long texts, each exceeding a million words, along with corresponding outlines.
\item Extensive experiments were conducted to validate the effectiveness of the proposed model, and the results show that our method achieves state-of-the-art performance.
\end{itemize}

\section{Related Works}
To the best of our knowledge, the tasks most closely related to outline generation for ultra-long fictional texts are Named Entity Recognition (NER), storyline generation, and outline generation, all of which have been extensively studied over the past few decades.

\subsection{NER}
Named Entity Recognition (NER) is a classical problem in natural language processing, aimed at automatically extracting named entities and their relationships from documents. In our research, NER is used to establish chapter node features. Early work on relationship extraction from text sources employed rules and patterns (e.g., ~\cite{ner1,ner2}). Open Information Extraction (Open IE) methods~\cite{ner3,ner4} use linguistic cues to infer patterns and triplets collectively, but they lack appropriate SPO (Subject-Predicate-Object) parameter normalization. With the recent advancements in pre-trained language models such as BERT, as well as ElMo, GPT-3, T-5, and others, the best current NER methods leverage these models for representation learning~\cite{bert-wwm,bestNer,bestNer2}.

\subsection{Storyline generation}
Storyline generation aims to summarize the development of certain events and understand how they evolve over time. \citet{story1} formalized different types of sub-events into local and global aspects. Several studies have used Bayesian networks for storyline detection~\cite{story2,story3}. \citet{story4} first obtained relevant tweets, and then generated story lines through graph optimization to extract the story plot of events. In \cite{story5}, an evolutionary hierarchy Dirichlet process was proposed to capture the theme evolution pattern in the plot summary. The current story line extraction focuses more on multi-modal data, such as \citet{story6} generating video story lines through structured story lines.

\subsection{Plot division and outline generation}
Early work primarily used unsupervised methods~\cite{seg1,seg2} for topic segmentation. As large-scale thematic structure corpora were developed, supervised methods gradually became mainstream, such as sequential labeling models~\cite{seg3,seg4}. Only a few studies have focused on Chinese topic segmentation tasks using sequential labeling models following English methodologies~\cite{chiseg1,chiseg2} or local classification models~\cite{chiseg3} to predict topic boundaries.

Most of the existing outline generation methods~\cite{pr_outline1,outline2} are for small and medium text, and more attention is paid to English. In our work, we introduce methods related to named entity recognition, title generation, and storyline generation; however, there are some notable differences. First, the NER task can only identify named entities and their relationships but cannot systematically construct graph data. Second, traditional plot segmentation and outline generation tasks output single document-level titles with coarse-grained semantics, whereas our outline generation (OG) task outputs a sequence of plot-level titles with fine-grained semantics and contextual identification. Lastly, storyline generation is based on multiple sub-events along a timeline, whereas the OG task focuses on multiple sections. Therefore, most of the existing methods for these related tasks may not be directly applicable to the OG task.

\section{Method}
For the overall architecture of our method, we divide it into five steps. These steps will be discussed in detail in this chapter.

\begin{figure*}[t]
	\centering
	\includegraphics[width=1\linewidth]{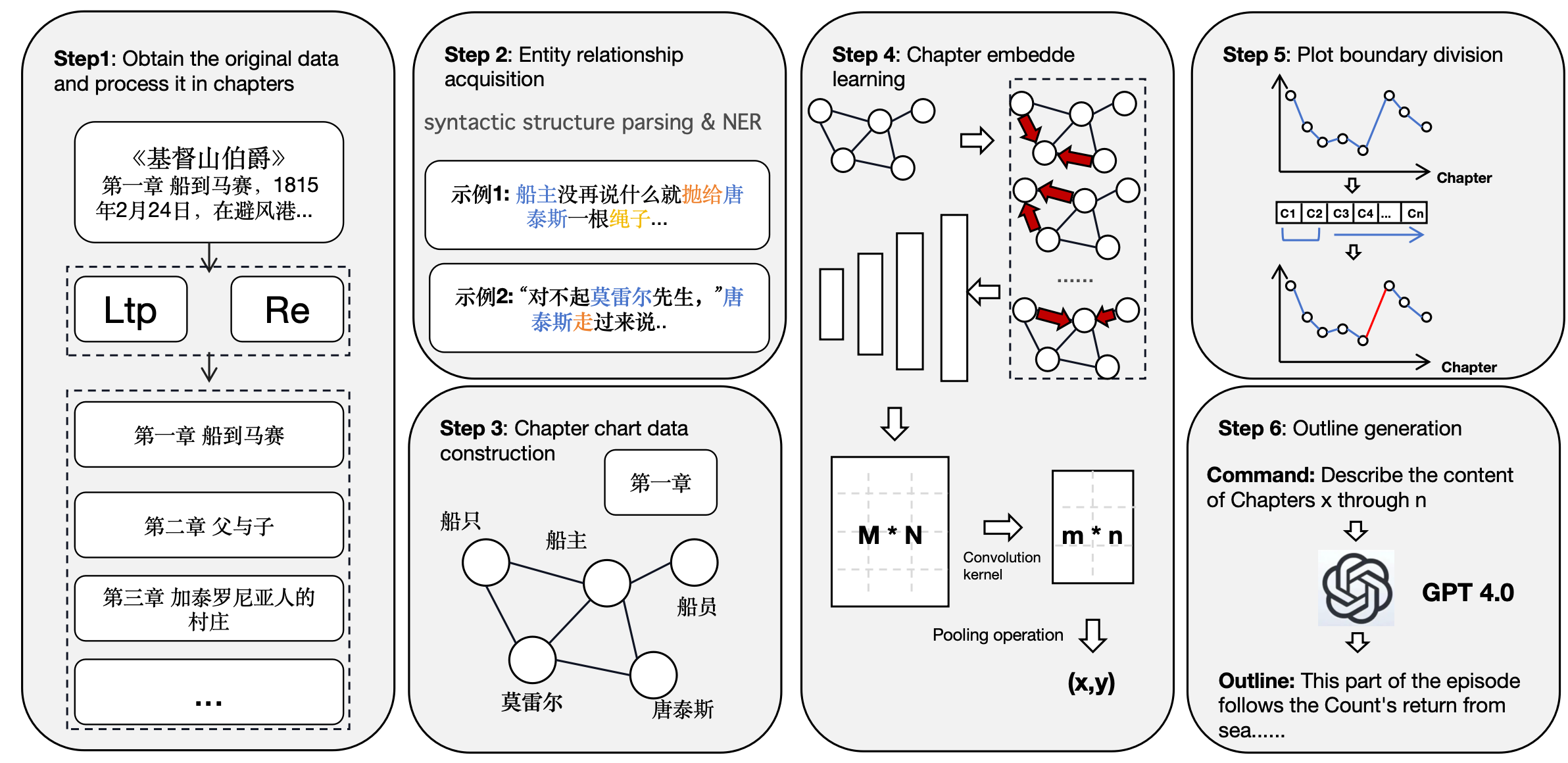}
	\caption{Multimodal
nonparametric clustering algorithm via Monte Carlo method and variational autoencoder diagram.}
	\label{flowchat}%文中引用该图片代号
\end{figure*}

\subsection{The text is processed in chapters}
For ultra-long fictional texts, it is necessary to construct an overall outline based on chapter information. Therefore, the first step is to divide the entire text according to its chapters. We use a regular expression method to match chapter titles, where each chapter begins with a label such as "Chapter X." This chapter title label allows for an effective segmentation of the entire text. In the context of Chinese, since Chinese characters are not formed by alphabet-like symbols, word tokenizers do not break words into smaller segments. Instead, they use Traditional Chinese Word Segmentation (CWS) tools to split the text into several words. This allows for the application of whole-word masking in Chinese, masking entire words instead of individual characters. To implement this functionality, the original whole-word masking code must be strictly followed, without altering other components such as the percentage of word masking. The LTP 
 tool is used for Chinese word segmentation to identify word boundaries. Similarly, whole-word masking can be applied to RoBERTa without using the NSP task.

\subsection{Construction of chapter level node eigenvector and adjacency matrix}

In text feature extraction, methods such as term frequency are often used, where important entity words represent the overall text features~\cite{proof_ner1,proof_ner2}. For each chapter of the text, we need to construct graph data to represent the content of the chapter. To do this, we first select chapter nodes and construct the feature vectors and adjacency matrices for these nodes. 

Therefore, for chapter-level text data, we utilize the LTP tool~\cite{ltp} for processing. Its core functionalities include word segmentation, part-of-speech tagging, named entity recognition (NER), and syntactic dependency analysis, among other sub-tasks. Similar to THULAC~\cite{THU}, LTP is also based on a structured perceptron (SP) and uses the maximum entropy principle to model the scoring function of the label sequence \( Y \) given the input sequence \( X \).
\begin{equation}
\label{L_t}
S(Y,X)=\sum_s\alpha_s\theta_s(Y,X)
\end{equation}

Here, $\theta_s$ represents the local feature function. The Chinese word segmentation problem is equivalent to solving for the label sequence \( Y \) corresponding to the score function, given the input \( X \).

Through the NER task, we can extract named entities. Using LTP for part-of-speech tagging, we select entities with noun tags as chapter nodes, which include key plot information such as the main characters, locations, and items within the chapter. After obtaining the nodes, the relationships between the entity nodes can be determined based on syntactic dependency information, allowing us to construct the adjacency matrix for the nodes within the chapter. $E(x,y) \in \{0,1\}$

To construct node features, we consider using the tf-idf matrix to highlight both node and chapter features. Therefore, we propose a new method for constructing chapter node features, which includes important attributes such as the entity name of the node, the chapter number, and the tf-idf value of the entity node.

For obtaining the vector for the entity name of a node, we consider using the BERT-WMM model~\cite{bert-wwm}. For the tf-idf[] values of entity nodes, we first filter the top 10 tf-idf values within the chapter, and then assign these tf-idf values to the chapter nodes. Each entity node corresponds to its respective tf-idf matrix value. If an entity node has one of the top 10 tf-idf values, the value is appended; otherwise, the corresponding matrix value is set to 0. This results in a 10-dimensional tf-idf value matrix, where each row represents a feature vector for an entity node.
Finally, the number of chapters is combined with the above features.

\subsection{Chapter deep embedding}

For learning chapter features, we have constructed an unsupervised learning model, a graph autoencoder based on GAT~\cite{gat} layers, to extract chapter features by learning node features and the adjacency matrix. The core idea is to use GAT to focus on each node’s neighbors in order to learn the hidden representation of the current node. At the same time, the AE model captures the feature vector attribute \( x_i \) of node \( v_i \). The most straightforward strategy for processing the neighbors of a node is to aggregate the node's representation equally with all of its neighbors. However, the importance of different neighboring nodes varies, which results in different weights being assigned to them. Based on the multi-head attention mechanism, the GAT network effectively strengthens the weights of important neighboring nodes while diminishing the weights of irrelevant ones. The computation of the hidden representation of the current node \( v_i \) is as follows:
\begin{equation}
Z^{l}_i = \sigma(\sum_{j\in N_i}\alpha_{ij}WZ^{l-1}_i)
\label{Z^{l}_i}
\end{equation}

$\alpha_{ij}$ represents the attention factor, signifying the importance of neighbor node $v_j$ to node $v_i$, and $\sigma$ denotes a nonlinear function. $W$ is a hyperparameter. $Z^{l}_i$ corresponds to the output representation of node $v_i$, and $N_i$ refers to the neighbors of node $v_i$. The subsequent step involves assessing the significance of neighbor node $v_j$ with consideration for both attribute value and topological distance, ultimately determining the attention coefficient $\alpha_{ij}$.

In terms of node attribute values, the $\alpha_{ij}$ can be expressed as a single-layer feedforward neural network for ${\mathop{x_i}\limits ^{\rightarrow}}$ and ${\mathop{x_j}\limits ^{\rightarrow}}$ in series with a weight vector of $\mathop{a}\limits ^{\rightarrow}\in \mathbb{R}$.
\begin{equation}
d_{ij}=a(W{\mathop{x_i}\limits ^{\rightarrow}},W{\mathop{x_j}\limits ^{\rightarrow}})
\end{equation}

Note that the coefficients are usually normalized between all neighbors $v_j \in N_i$ with the softmax function, making them easy to compare between nodes.
\begin{equation}
\alpha_{ij}=softmax_j(d_{ij})=\frac{exp(d_{ij})}{\sum_{k\in N_i}exp(d_{ik})}
\end{equation}

Following this, the representation of the current target node by its neighboring nodes in terms of topology becomes essential. GAT, in its original form, concentrates solely on the 1-hop neighboring nodes (first-order) of the current node \cite{gat}. However, given the intricate structural relationships within graphs, there arises a need for higher-order neighboring node relationships. To address this, we stack $n$ layers of GAT, enabling the current node $v_i$ to retain information about higher-order neighbors. This can be formulated as:
\begin{equation}
H=\sum_{j\in N_i}\alpha_{ij}v_j=\sum_{j\in N_i}\alpha_{ij}(x_1,x_2......,x_n),x_i \in \mathbb{R}^{N \times d}
\end{equation}

We choose to reconstruct the graph structure as part of the decoder, which uses the $sigmoid$ function to map $(-\infty,+\infty)$ to the probability space. We minimize the reconstruction error by measuring the difference between $A_i$ and $A'_i$, where $A'_i$ is the reconstructed structure matrix of the graph.
\begin{equation}
L_r=\sum^n_{i=1}loss(A_i,A'_i),
A'_i=sigmoid(Z^TZ)
\label{A_I}
\end{equation}

Then we reduce the learned deep embeddings to 2-dimensional data on the feature space through a pooling layer to obtain chapter feature embeddings $Z$.

\subsection{Plot boundary division}

We propose a new method based on the principles of path dependence and Markov chains~\cite{macv}, referred to as DMc. This allows us to further predict plot boundaries based on the learned chapter features. This step involves dividing the segmented ultra-long document \( c_1, c_2, c_3, \dots, c_n \) into multiple consecutive parts \( \{s_1, s_2, \dots, s_N\} \) by predicting the section boundary labels \( \{l_1, l_2, \dots, l_M\} \), where \( l_M \in \{0,1\} \). If \( l_m = 0 \), then \( c_n \) is a chapter within a plot boundary, and the section prediction continues. If \( l_m = 1 \), then \( c_n \) is the last chapter of a plot segment, and an appropriate title should be generated. Some literature points out that paragraphs are coherent units of information; we consider chapters as sequences of coherent paragraphs, and coherence modeling is inherently non-trivial. The properties of Markov chains can help address consistency between contexts and identify paragraph boundaries. However, it is undeniable that plot boundaries are still related to content mentioned in the previous chapters. Therefore, we introduce the principle of path dependence and construct an operator mechanism to predict plot boundaries in ultra-long texts.

The operator specifically extends the process of examining the previous chapter of the target chapter to considering the previous \( \alpha \) chapters in order to determine whether the chapter is a plot boundary. As shown in Figure 2, we use the hidden representations of the current chapter \( c_m \), the previous \( \alpha \) chapters \( c_{m-\alpha} \), and the next chapter \( c_{m+1} \) to predict the segment boundary label \( l_m \). We set a threshold \( s_t \), determined by the EU (Embedding Unit) of the previous \( \alpha \) chapters, with the following formula:

\begin{equation}
s_t = \beta \cdot mean(EU)
\end{equation}

where \( \beta \) is the learning parameter. If \( EU(c_{m+1}) > s_t \), then the current chapter \( c_m \) is considered a plot boundary. Otherwise, it is considered to be part of the same plot. Next, we set a safety distance \( d_d \), which represents the minimum number of chapters that we consider as part of the ongoing plot to save computational resources. Therefore, the operator will continue searching for the plot boundary after \( c_{m+d_d} \).

\subsection{Outline of stories based on LLM}
After the plot content is summarized, chapter boundaries and related instructions are input into the large model to obtain the overall text outline.

\section{Experiments}
\subsection{Dataset}
In this experiment, a Chinese data set containing 31 ultra-long texts for outline generation is constructed for us. The topics covered include adventure, fantasy, fairy, biography, classics five categories. The average text size is 2234.63kb and contains an average of 580.8 chapters.As Table~\ref{datasettable}

\begin{table}[t]
  \caption{Information of dataset}
  \label{datasettable}
  \centering
  \begin{tabular}{llll}
  \toprule
    Dataset     & Episodes     & Chapters & Size\\
  \midrule
    20News & 2.7  & 413.2   & 211.53kb    \\
    FG & 3.8  & 310   & 134.37kb       \\
    Ours & 4.5  & 580.8   & 2234.63kb    \\
  \bottomrule
  \end{tabular}
\end{table}

\subsection{Baseline}
GPT 3.5:  GPT-3.5 is a large language model based on the GPT-3.5 architecture that utilizes a network of transformers to perform various tasks such as dialogue, text completion, and language translation.

GPT 4.0: A new generation of GPT models, exceeding 3.5 in both size and performance.

Lama7b: Excellent open source large model based on transformer

\subsection{Evaluation Metric}
To evaluate the performance of the proposed method, we used the accuracy, recall rate, and F-score commonly used to evaluate information extraction systems. Accuracy is calculated based on the following conditions: whether the division of plot boundaries is correct. In addition, we also evaluated the readability of the generated outline from two different evaluation indexes. These are CheckEval Framework and Kendall tau.

\subsection{Result}
\begin{figure}[t]
\centering
\subfloat[Various CheckEval aspects results]{
		\includegraphics[width=0.25\textwidth]{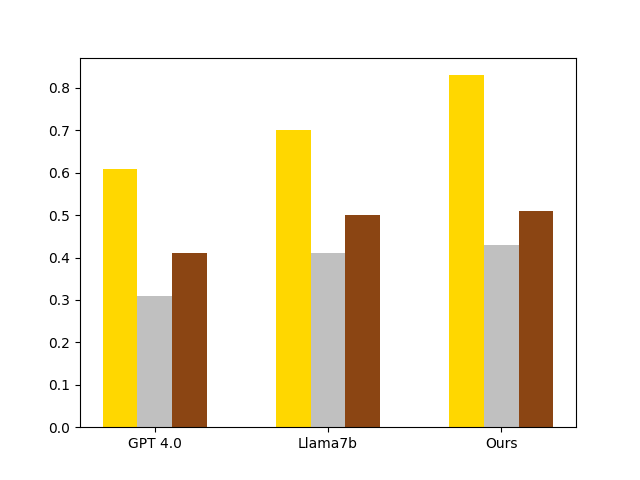}}
\subfloat[Kendall tau correlations]{
		\includegraphics[width=0.25\textwidth]{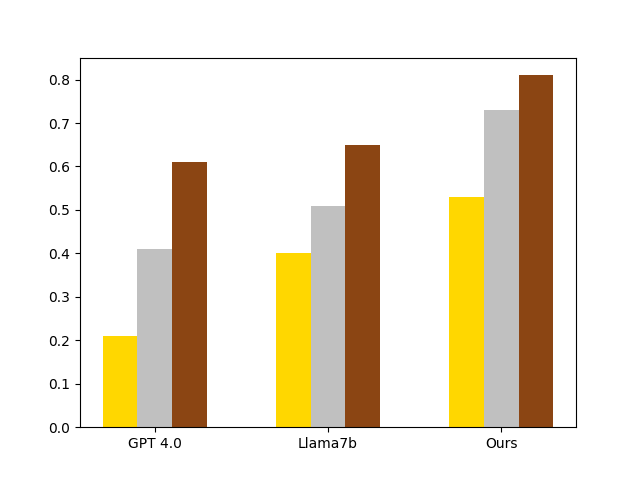}}
\caption{The variation of ACC and NMI on each dataset for different values of $\alpha$. And plot of results of the number of clusters on the MNIST datasets judged using other parameters.}
\label{figtau}
\end{figure}

We evaluate the model's performance from two aspects: boundary prediction accuracy and the readability of the generated outline. First, we test boundary prediction accuracy on both our constructed dataset and two publicly available datasets, as shown in Table~\ref{acc}. Additionally, the readability of the generated outline is tested using the CheckEval framework and the Kendall Tau correlation, as presented in Fig~\ref{figtau}.

Furthermore, we conduct detailed experiments on each ultra-long text in our constructed dataset, with the relevant experimental data provided in Appendix A. This includes tests for plot boundary prediction accuracy and readability analysis for each book. We also provide several detailed outline examples to demonstrate the readability of our model.

\begin{table}[t]
  \caption{ACC of result(\%)}
  \label{acc}
  \centering
  \begin{tabular}{llll}
  \toprule
    Dataset     & 20NEWS     & FG & Ours\\
  \midrule
    gpt3.5 & 27.0  & 14.3   & 20.1    \\
    gpt4.0 & 37.1  & 11.3   & 26.1         \\
    Llama7b & 17.0  & 34.3   & 37.3    \\
    Ours & 57.1  & 44.0   & 87.5   \\
  \bottomrule
  \end{tabular}
\end{table}

From the above experimental results, it is obvious that our method is better in predicting the accuracy of plot boundaries. This makes our generation outline more accurate. In addition, two index tests on outline readability show that our generated outline is more readable.

\section{Conclusion}
In this paper, we propose a method based on plot segmentation to guide large models in generating better outlines for ultra-long texts. First, the chapter graph data effectively captures chapter feature information. Based on the chapter embeddings learned by the GAT, we use an improved Markov chain to divide the plot boundaries. Finally, the large model accurately generates the plot content for each boundary, which is then aggregated into the outline. When compared with several deep learning models and large models, our performance achieves optimal results. Future research will focus on how to integrate the preceding steps into large models.
\bibliography{references}
\bibliographystyle{plainnat}

\end{document}